\PassOptionsToPackage{unicode}{hyperref}
\PassOptionsToPackage{hyphens}{url}
\PassOptionsToPackage{dvipsnames,svgnames,x11names}{xcolor}
\documentclass[
  10pt,
  letterpaper,
]{article}
\usepackage{xcolor}
\usepackage[letterpaper,width=5.5in,height=9in,centering]{geometry}
\usepackage{amsmath,amssymb}
\usepackage{iftex}
\ifPDFTeX
  \usepackage[T1]{fontenc}
  \usepackage[utf8]{inputenc}
  \usepackage{textcomp} %
\else %
  \usepackage{unicode-math} %
  \defaultfontfeatures{Scale=MatchLowercase}
  \defaultfontfeatures[\rmfamily]{Ligatures=TeX,Scale=1}
\fi
\usepackage{lmodern}
\ifPDFTeX\else
\fi
\IfFileExists{upquote.sty}{\usepackage{upquote}}{}
\IfFileExists{microtype.sty}{%
  \usepackage[]{microtype}
  \UseMicrotypeSet[protrusion]{basicmath} %
}{}
\usepackage{setspace}
\makeatletter
\@ifundefined{KOMAClassName}{%
  \IfFileExists{parskip.sty}{%
    \usepackage{parskip}
  }{%
    \setlength{\parindent}{0pt}
    \setlength{\parskip}{6pt plus 2pt minus 1pt}}
}{%
  \KOMAoptions{parskip=half}}
\makeatother
\usepackage{graphicx}
\makeatletter
\newsavebox\pandoc@box
\newcommand*\pandocbounded[1]{%
  \sbox\pandoc@box{#1}%
  \Gscale@div\@tempa{\textheight}{\dimexpr\ht\pandoc@box+\dp\pandoc@box\relax}%
  \Gscale@div\@tempb{\linewidth}{\wd\pandoc@box}%
  \ifdim\@tempb\p@<\@tempa\p@\let\@tempa\@tempb\fi%
  \ifdim\@tempa\p@<\p@\scalebox{\@tempa}{\usebox\pandoc@box}%
  \else\usebox{\pandoc@box}%
  \fi%
}
\def\fps@figure{htbp}
\makeatother
\providecommand{\tightlist}{%
  \setlength{\itemsep}{0pt}\setlength{\parskip}{0pt}}
\usepackage[style=apa,backend=biber,natbib=true]{biblatex}
\usepackage{amsmath}
\usepackage{amssymb}
\usepackage{amsthm}
\usepackage{mathtools}
\usepackage{bm}

\usepackage{graphicx}
\usepackage{caption}
\usepackage{subcaption}
\usepackage{float}
\usepackage{wrapfig}

\usepackage{setspace}
\usepackage{indentfirst}

\usepackage{microtype}

\usepackage{booktabs}
\usepackage{longtable}
\usepackage{array}
\usepackage{multirow}
\usepackage{threeparttable}

\usepackage{hyperref}
\hypersetup{
  colorlinks=true,
  linkcolor=black,
  citecolor=blue,
  urlcolor=blue,
  pdfborder={0 0 0}
}

\usepackage{booktabs}
\usepackage{longtable}
\usepackage{array}
\usepackage{multirow}
\usepackage{wrapfig}
\usepackage{float}
\usepackage{colortbl}
\usepackage{pdflscape}
\usepackage{tabu}
\usepackage{threeparttable}
\usepackage{threeparttablex}
\usepackage[normalem]{ulem}
\usepackage{makecell}
\usepackage{xcolor}
\usepackage{bookmark}
\IfFileExists{xurl.sty}{\usepackage{xurl}}{} %
\hypersetup{
  pdftitle={Semantic Alignment of AI Models: Concept Collapse, Checkpoint Dynamics, and Cross-Lingual Transfer},
  pdfauthor={Tyler Ashoff; Jordan Rodu},
  colorlinks=true,
  linkcolor={blue},
  filecolor={Maroon},
  citecolor={blue},
  urlcolor={blue},
  pdfcreator={LaTeX via pandoc}}

\title{Semantic Alignment of AI Models: Concept Collapse, Checkpoint
Dynamics, and Cross-Lingual Transfer}
\author{Tyler
Ashoff\thanks{Department of Statistics, University of Virginia. Correspondence: tlashoff@gmail.com. This work is derived in part from the first author's doctoral dissertation \parencite{ashoff2026dissertation}.} \and Jordan
Rodu\thanks{Department of Statistics, University of Virginia}}
\date{August 2026}

\begin{document}
\maketitle

\setstretch{1}
Language model benchmarking is a difficult task. Outcome reasoning alone
does not test the model's conceptualization of language and popular
open-source benchmarks are quickly saturated or ingested as training
data. It is important to test the model's output, but augmenting these
tests by characterizing semantic structure gives more insight to how
models relate abstract concepts. However, the high dimensional embedding
spaces are not easy to interpret. This work demonstrates how topological
methods can be used to rigorously compare these spaces to low
dimensional and interpretable baselines like ontologies and curated
knowledge graphs. These multi-modal alignment tests make it possible to
track model adaptations and test phrase understanding across multiple
languages.\footnote{Implementation available at \href{https://github.com/tylerashoff/persiscope}{github.com/tylerashoff/persiscope} and on PyPI as \texttt{persiscope}.}

\begin{center}\rule{0.5\linewidth}{0.5pt}\end{center}

\section{Introduction}\label{introduction}

\subsection{Motivation}\label{motivation}

Large language models (LLMs) are difficult to evaluate rigorously.
Outcome based benchmarks are a popular method for grading general
competence, but the tests are not exhaustive due to their limited scale
compared with the vast output space of the models. Red teaming and human
evaluation help fill the gaps in these tests, but they are expensive and
labor intensive. Popular models benefit from increased attention by
surfacing more failure modes and edge cases, but even the attention of
many users will leave vulnerabilities unfound. This problem is only
exacerbated for less popular or more niche models, their outcome space
remains extremely broad but there are fewer users to test the models.

This work demonstrates how a topological alignment approach effectively
compares models' semantic structure to curated baselines. Rather than
being restricted by string output or multiple choice methods like many
benchmarks or red-team approaches, this method leverages the models'
embedding space to test coherence. The advantage of these techniques
when test sets are saturated is demonstrated in the companion paper
\parencite{persistent-convolution} and extended in this work. We show
that the method effectively investigates training dynamics,
cross-lingual alignment, and the manipulation of LLM behavior.

\subsection{Existing Work}\label{existing-work}

AI alignment focuses on how to ensure human preferences are learned by
and evident in the output of models. Many papers have outlined the
common failure modes of AI. These include unintended consequences caused
by domain shifts or model extrapolation \autocite{ai-safety-problems},
adversarial inputs \autocite{gibberish} from users with prompt injection
\autocite{prompt-injection} and jailbreaking \autocite{jailbreak}
\autocite{autodan}, or more insidious model behavior like reward hacking
\autocite{spec-gaming} and sleeper agent models
\autocite{sleeper-agents}. Many solutions are proposed to these
problems, but Reinforcement Learning with Human Feedback (RLHF)
\autocite{rlhf} has become an industry standard for model training.
Other approaches like Anthropic's Constitutional AI \autocite{const-ai},
and what they internally call the ``Soul Doc''\autocite{soul-doc}, try
to impart values to the model with ``a detailed description of
{[}\ldots{]} the kind of entity we would like Claude to
be''\autocite{soul-doc-ann}. To provide oversight of large models at a
scale where RLHF may fail, ``weak-to-strong
generalization''\autocite{weak-to-strong} is employed, where a smaller
model is trained to supervise a larger model by generalizing human
interpretable safety instructions. However, other work also tries to
understand if these setups can be trusted, or if LLMs can effectively
fake alignment \autocite{alingment-faking}.

The mechanistic interpretability field is also interested in model
behavior, but focuses on finding ``circuits''
\autocite{transformer-circuits}, or subgraphs of a model's network, that
correlate with certain behaviors. This interest in circuits has driven
work on monosemanticity \autocite{monosemanticity}, sparse autoencoders
(SAEs) \autocite{sae}, and transcoders \autocite{circuit-tracing} that
try to isolate and identify features that activate for specific
concepts.

Like alignment and mechanistic interpretability, semantic structure
research tries to understand how to elicit certain model behaviors. But,
rather than shaping it with human feedback, or looking for circuits, the
work tries to understand the latent representations of these behaviors.
The hope is that vectors can be found in the high-dimensional latent
space that correspond to values or concepts in a model. The classic
example shows how vectorization of words may be combined to morph one
concept into another: ``\(King - Man + Woman = Queen\)''
\autocite{word2vec}. Finding and manipulating these vectors related to
high-level concepts like ``honesty'', ``power-seeking'', or
``happiness'' is called Representation Engineering (RepE)
\autocite{rep-eng}.

This informed Bidirectional Encoder Representations from Transformers
(BERT) \autocite{bert} which introduced contextual embeddings and
provided the structure for modern transformer models. Studying the
emergent world representations \autocite{othello} of a model is an
active field of research, including the representation of categorical
concepts as polytopes \autocite{polytopes} and truth vectors
\autocite{geometry-of-truth}. Work related to the Manifold Hypothesis in
representation learning \autocite{manifold-hypothesis} suggests that
complex concepts may be associated with linear directions
\autocite{linear-directions} in the model's representation space. Other
work has leveraged this hypothesis to show that LLMs may have a coherent
world view by demonstrating a spatio-temporal \autocite{spatio-temporal}
understanding of places and historical figures.

\subsection{Limitations of Benchmarking and Test
Sets}\label{limitations-of-benchmarking-and-test-sets}

Language models are often graded using benchmarks and open source human
evaluation networks like the ``Chatbot
Arena''\autocite{chatbot-arena-24}. These methods are primarily task
based; i.e., they are designed with a set of tasks each with a set of
examples which are graded using various measures of correctness. Three
popular examples of these benchmarks are MMLU, HELM, and BIG-Bench.
These tasks often take the form of multiple-choice questions generated
by experts or, in the case of MMLU, pulled directly from standardized
tests. Often these models have been trained on most of the Internet,
which includes published standardized tests. Even a stochastic parrot
would excel at the recall-based tests that are used in the popular
benchmarks.

Prediction focused tests are an important type of evaluation since,
ultimately, the predicted word sequences are the product of these
models. However, embedding spaces offer a richer, numerical
representation of these outputs. The embedded data reveals how a model
builds relationships between data points and broader concepts, if the
data can be grouped into classes.

\begin{table}[H]
\centering\centering
\caption{\label{tab:pop-bench-tab}\label{tab:pop-bench}Popular Benchmarks}
\centering
\resizebox{\ifdim\width>\linewidth\linewidth\else\width\fi}{!}{
\begin{tabular}[t]{>{\centering\arraybackslash}p{7em}>{\centering\arraybackslash}p{10em}>{\centering\arraybackslash}p{20em}}
\toprule
Benchmark & Focus & Description\\
\midrule
\cellcolor{gray!10}{MMLU} & \textbf{\cellcolor{gray!10}{Standardization}} & \cellcolor{gray!10}{Built directly from multiple-choice standardized tests, but becoming saturated}\\
\addlinespace\addlinespace
HELM & \textbf{Safety} & Developed around scenarios and question answering, more niche\\
\addlinespace\addlinespace
\cellcolor{gray!10}{BIG-Bench} & \textbf{\cellcolor{gray!10}{Broad Task Set}} & \cellcolor{gray!10}{Continually growing set of tasks designed to avoid saturation}\\
\addlinespace\addlinespace
Chatbot Arena & \textbf{Human Preference} & Humans pick between outputs from two LLMs for the same prompt\\
\bottomrule
\end{tabular}}
\end{table}

The Measuring Massive Multitask Language Understanding (MMLU)
\autocite{hendrycks2021measuring} benchmark is ``a massive multitask
test consisting of multiple-choice questions from various branches of
knowledge''. The questions, 15908 across 57 tasks, ``were manually
collected by graduate and undergraduate students from freely available
sources online''. In spirit it is a multiple-choice standardized test
for language models, in fact many of the questions come directly from
standardized tests. It seems intuitive to grade language models like we
grade students and it has been a standard metric, but it has become
saturated. Additionally, since it is only a multiple-choice test,
sensitivity analysis is limited by the number of questions and the
trickiness of the answers.

The Holistic Evaluation of Language Models (HELM)
\autocite{liang2023holistic} evaluation framework is described as a
``living benchmark''. The benchmark's core evaluation set is 16
scenarios and 7 metrics designed to evaluate model performance in
multiple domains, and continues to add more multilingual and multimodal
scenarios. It is impressive work covering many of the most powerful
models and evaluating them across many dimensions. Their measure of
Robustness is of primary interest to this work.

HELM ``measure{[}s{]} the robustness of different models by evaluating
them on transformations'' of the data sets and ``measure{[}s{]} the
worst-case performance of a model across these transformations''. These
transformations are broken into two types: ``\emph{invariance and
equivariance}''. Invariance tests ``natural and relatively mild''
changes like typos while equivariance tests ``semantics-altering
perturbations''. However, the authors note that due to limitations in
generating realistic semantic changes and specifying expected output
changes for long output sequences, the use of these measures is
relatively limited. The authors also point out, these metrics only
capture the ``\emph{local} robustness of a model'' and generating enough
changes to approximate the true worst-case performance is ``not feasible
in this evaluation''. These issues motivate the need for a model
characterization that captures global and local performance.

The Beyond the Imitation Game Benchmark
(BIG-bench)\autocite{srivastava2023imitation} consists of a set of 204
language tasks, but is constantly growing as developers continue to add
to the project and has been updated to BIG-Bench Hard
\autocite{big-bench-hard-23} and Extra Hard
\autocite{big-bench-extra-hard-25} variants to avoid saturation. On the
high end these tasks have on the order of \(10^6\) examples while most
tasks have around \(10^2\)-\(10^3\) examples. BIG-Bench provides a set
of available metrics for developers to choose from. Some of the metric
options are presented in Table \ref{tab:bb-mets}. Exact string matching
is the most restrictive of all the metrics. Bleu \autocite{bleu2002} and
ROUGE \autocite{lin2004rouge} allow for some flexibility in word choice
and word arrangement. Bleurt \autocite{sellam2020bleurt} uses a
secondary model to grade the output of the first, both multiple-choice
methods rely on the available choices and the calibration version
incorporates the model's own internal confidence. Yet none of them offer
a natural model characterization measuring the sensitivity of these
results relies on the scale of the test set.

Many types of benchmarks exists for different types of models. Agentic
reasoning tests, like ARC-AGI \autocite{arc-agi}, may ask the system to
solve ambiguous interactive puzzles for example. Despite, their
complexity and novelty, these benchmarks still suffer from the same
limitations. They are restricted by the variety and relevance of
individual question sets. They do not directly measure how the model
conceptualizes a domain.

The insufficiency of these metrics can be seen mostly clearly in the
continued popularity of Chatbot Arena and its variants. Chatbot Arena
shows a human evaluator the output from two models using the same prompt
and asks the human to choose the better response. The cumulative result
of many human evaluations is a crowd-sourced model ranking. This ranking
fills the gap in other frameworks by capturing some ill-defined human
alignment score, but it does not try to make measurement explicable or
rigorous. If models are to be trusted in unfamiliar situations,
understanding how the model conceptualizes the world is more important
than if the model was right or wrong on an inexhaustive test set.

\begin{table}[H]
\centering
\caption{\label{tab:bigbench-mets}\label{tab:bb-mets}Sample of BIG-Bench available metrics.}
\centering
\begin{tabular}[t]{>{\centering\arraybackslash}p{8em}>{\centering\arraybackslash}p{8em}>{\centering\arraybackslash}p{20em}}
\toprule
Metric & Type & Description\\
\midrule
\cellcolor{gray!10}{Bleu} & \textbf{\cellcolor{gray!10}{Text-to-text}} & \cellcolor{gray!10}{Weighted average of subsequence alignment between output and reference word sequences.}\\
\addlinespace\addlinespace
Bleurt & \textbf{Text-to-text} & Relies on another model, BERT, to simulate human judgment.\\
\addlinespace\addlinespace
\cellcolor{gray!10}{ROUGE} & \textbf{\cellcolor{gray!10}{Text-to-text}} & \cellcolor{gray!10}{Counts the overlapping subsequences between output and reference word sequences.}\\
\addlinespace\addlinespace
exact.str.match & \textbf{Text-to-text} & Binary indicating if the strings are identical.\\
\addlinespace\addlinespace
\cellcolor{gray!10}{multiple.choice grade} & \textbf{\cellcolor{gray!10}{Multiple-choice}} & \cellcolor{gray!10}{Observed probability of correct choice.}\\
\addlinespace\addlinespace
calibration multiple.choice brier.score & \textbf{Multiple-choice} & Observed probability of correct choice compared to the model's internal confidence score.\\
\bottomrule
\end{tabular}
\end{table}

\subsubsection{Task-based Testing in the Real
World}\label{task-based-testing-in-the-real-world}

These benchmarks curate a large test set and evaluate the models' output
on correctness by combining scores on different tasks into a final
grade. These evaluations fundamentally rely on an ever-growing and
community-sourced test set. However, by the benchmarks' own admission,
it is not clear that generating an exhaustive test set is even possible
given the enormous size of the output space of these models. Their scale
is impressive, but \emph{Webster's Third New International Dictionary}
\parencite*{websters} has about \(470,000\) entries and on the order of
\(10^{16}\) combinations for 3-grams. Of course, many of these
combinations will be nonsensical, but the input to language models is
not restricted to proper grammar and most texts are longer than three
words. This estimate also does not take into account multilingual models
or subtokenization of words. Additionally, prompt injection and
jailbreaking can redirect prompt instructions or avoid built-in
guardrails respectively, so both coherent and nonsense input can lead to
undesirable output.

This overview has focused on language models, but the problems are
similar to the issues models face in other domains. Although the output
space is large, language is still a discrete problem limited in scope by
the number of characters in a given vocabulary. Image, audio, or
spectral data are continuous domains. While they may be discretized for
processing, their true input and output spaces are infinite.

Frontier labs like Anthropic are also facing these challenges while
evaluating models.

\begin{singlespace}
\begin{quote}
\textit{"For AI R\&D capabilities, we found that Claude Opus 4.6 has saturated most of our automated evaluations, meaning they no longer provide useful evidence for ruling out [safety measures. \dots] \\[1ex]
We may build more tasks to test capabilities in the long-horizon regime, but evaluations of model capabilities may also need to depend more on expert judgments[.]"}

\hfill -- Claude Opus 4.6 System Card, \textcite{opus46-system-card}
\end{quote}
\end{singlespace}

The paper introducing BIG-Bench starts with a prescient quote which
suggests the opacity and complexity of modern models is inherent to
competent learning machines.

\begin{singlespace}
\begin{quote}
\textit{"An important feature of a learning machine is that its teacher will often be very largely ignorant of quite what is going on inside."}

\hfill -- Alan \textcite{turing}
\end{quote}
\end{singlespace}

In some respects the value of these metrics lean on Turing's quote. If
we are left with only the output for a model, we can only check its
sample correctness. But we have access to the internal model embeddings,
and while they may not provide us with a complete understanding of the
system, we can still leverage the information the embeddings provide to
peer inside. Even if we remain ``very largely ignorant'' of a model's
inner workings, we do not need to pretend to be completely ignorant.

The main contributions of this work are:

\begin{itemize}
\tightlist
\item
  Induced concept collapse can be tracked as increasingly aggressive
  malicious adaptations are applied
\item
  Semantic structure alignment can be characterized during model
  training
\item
  Multilingual comparisons can be made using a method that captures
  semantic meaning more naturally than existing methods
\item
  Changes in a model's internal semantic understanding can be tracked
  across layers
\end{itemize}

\begin{center}\rule{0.5\linewidth}{0.5pt}\end{center}

\section{Testing Procedure}\label{testing-procedure}

A full development of the testing procedure is available in the
companion paper \parencite{persistent-convolution} and an outline is
provided here. The main test is a formal homogeneity test
\autocite{equal-dist-high-dim} for embedding spaces based on energy
statistics \autocite{you2022comparing} and topological representations
called persistence landscapes \autocite{bubenik2015statistical} and
persistence silhouettes \autocite{pers-silh}. These topological methods
result in stable representations of a model's semantic structure that
focus on the connectivity of the space rather than absolute scale, are
invariant to homomorphic transformations, and are agnostic to the
dimensionality of the embedding space. Regardless of the dimensionality
these techniques produce smooth two-dimensional curves. This allows the
method to characterize the semantic structure or concept-space
clustering more meaningfully than other graph matching or clustering
methods.

Table \ref{tab:alignment-test} walks through the process of generating
these representations from embeddings, running pairwise comparisons on
these representations, and testing statistical significance at a
Benjamini--Yekutieli \autocite{by-p-value} adjusted level for multiple
comparisons.

\begin{table}[H]
\centering
\begin{tabular}{r p{4.4in}}
\toprule
\multicolumn{2}{l}{\textbf{Input:} embeddings $E_1$, $E_2$ of the same inputs; bootstrap count $n$;} \\
\multicolumn{2}{l}{\hspace{3.2em} permutation count $B$; significance level $\alpha$} \\
\multicolumn{2}{l}{\textbf{Output:} energy statistic $T$; p-value $p$} \\
\midrule
1 & for $r = 1, \dots, n$: \\
2 & \quad $S_1, S_2 \leftarrow \text{subsample}(E_1),\ \text{subsample}(E_2)$ \\
3 & \quad $D_1, D_2 \leftarrow$ persistence diagrams of $S_1$, $S_2$ \\
4 & \quad transform $D_1$, $D_2$ (rotate and rescale; extra $H0$ rotation) \\
5 & \quad $\lambda_{1,r}, \lambda_{2,r} \leftarrow$ landscape (or silhouette) of $D_1$, $D_2$ \\
6 & $T \leftarrow$ energy statistic between $\{\lambda_1\}$ and $\{\lambda_2\}$ \\
7 & $\text{pool} \leftarrow \{\lambda_1\} \cup \{\lambda_2\}$ \\
8 & for $b = 1, \dots, B$: \\
9 & \quad randomly split pool into two sets of size $n$ \\
10 & \quad $T^{(b)} \leftarrow$ energy statistic between the permuted sets \\
11 & $p \leftarrow \bigl(1 + \#\{b : T \leq T^{(b)}\}\bigr) / (B+1)$ \\
12 & reject $H_0$ (equal semantic structure) at level $\alpha$ if $p \leq \alpha$ \\
\bottomrule
\end{tabular}
\caption{Topological alignment testing procedure}
\label{tab:alignment-test}
\end{table}

\begin{figure}[p]
\centering

\begin{center}\includegraphics[width=0.95\linewidth]{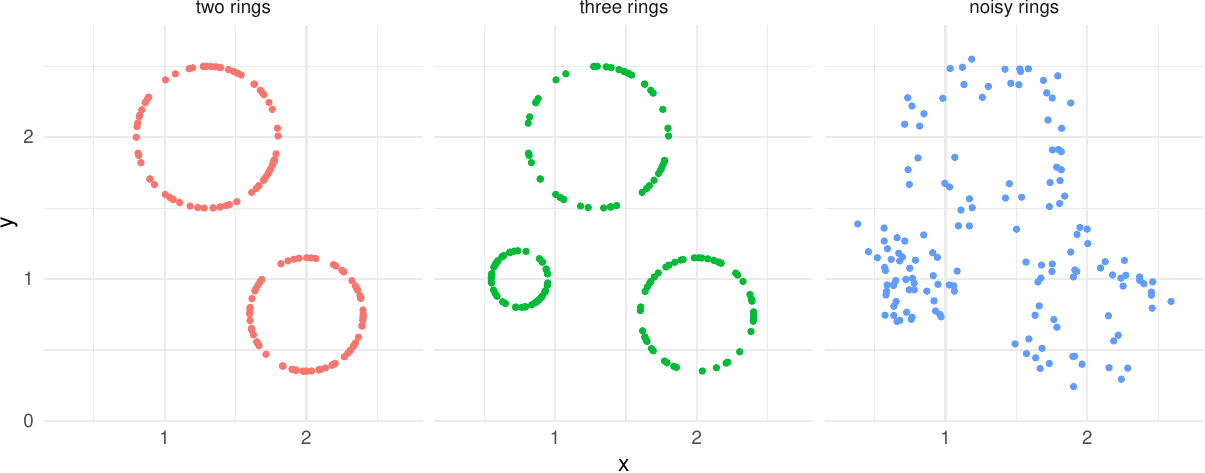} \end{center}
\caption{Three ring embeddings of 150 points each: two clean rings, three clean rings, and three noisy rings}
\label{fig:ring-embeddings}
\end{figure}

\begin{figure}[p]
\centering

\begin{center}\includegraphics[width=0.75\linewidth]{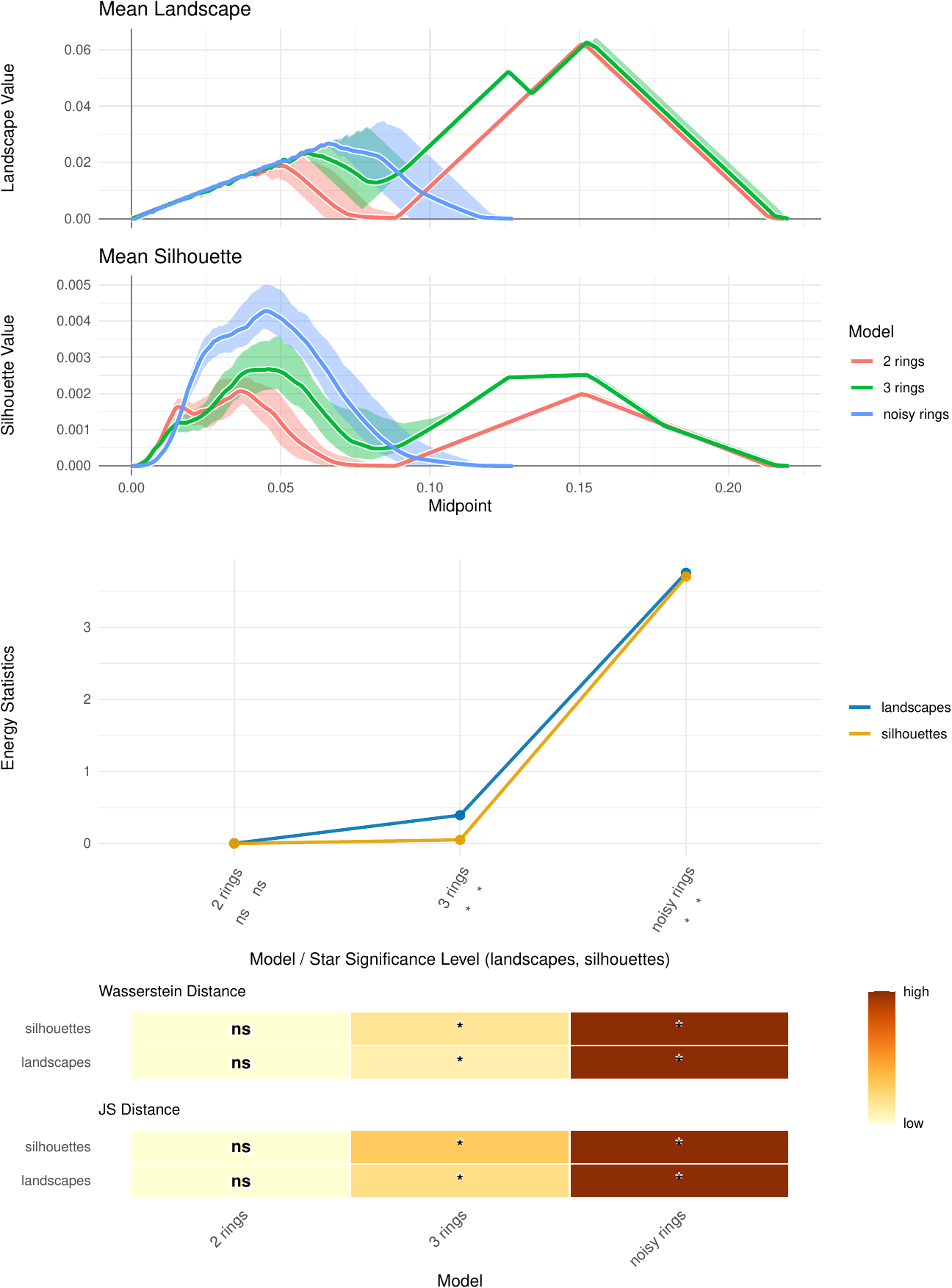} \end{center}
\caption{Ring Example Results: (top) Mean representations of three embeddings with 95\% point-wise confidence bounds. (middle) Line plot of test statistics comparing each embedding to the two ring baseline with p-values represented in star notation under the model labels. (bottom) Heatmaps with the Wasserstein distance and JS distance between three embeddings and the two ring baseline.}
\label{fig:energy-stat-heatmap-line-example}
\end{figure}

Figure \ref{fig:ring-embeddings} shows three embeddings of 150 points
each: two clean rings, three clean rings, and three noisy rings. Figure
\ref{fig:energy-stat-heatmap-line-example} shows the topological
representation of these embeddings and the tests associated with these
comparisons. The top two plots show the persistence landscapes and
persistence silhouettes of the three embeddings. Especially in the
landscapes, it is clear how the representations capture the clusters of
points in the peaks of the plots. Three peaks for three clean rings, two
peaks for two rings, and one peak as the noisy data blurs the underlying
structure. The lines are shown with 95\% pointwise confidence intervals.

The middle plot shows the test statistic when each embedding is compared
to the two ring baseline. As expected, the test statistic is 0 and not
significant when the two ring baseline is compared to itself. Then the
clean three ring embedding is statistically different from the two
rings, but more similar to the two ring embedding than the noisy three
rings.

On the bottom are extra curve similarity metrics. The same topological
representations are used and more standard similarity scores are shown
to demonstrate the consistency of results. The Wasserstein distance is a
natural metric for landscapes and the Jensen-Shannon Distance (JSD) is a
natural metric for silhouettes, the scores are normalized within each
representation. These metrics show similar values as the
energy-statistics but prove to be less discriminative.

\begin{center}\rule{0.5\linewidth}{0.5pt}\end{center}

\section{Concept Collapse via Language Model
Adaptation}\label{concept-collapse-via-language-model-adaptation}

The first result demonstrates how this method can detect subtle failure
modes. Increasingly adversarial LoRA adaptations are applied with the
goal of degrading the model's semantic structure, and the topological
tests are able to track this degradation effectively.

Language models typically have vocabularies consisting of a collection
of common words, subwords, single letters, and other prefixes or
suffixes of words. This diversity of combinations allows the model to
ingest any sequence of characters even if it is unfamiliar. How a model
breaks apart words is called tokenization strategy and it varies
significantly from model to model. However, if a word is subtokenized,
or broken apart, it may be represented by multiple embeddings rather
than a single vector. Many strategies aggregate multi-token sequences,
but the different strategies may be difficult to compare. That question
is avoided in this section, only examples where the word of interest
(WOI) is not subtokenized are retained. This results in 358 samples for
the OLMo model and 284 samples for the Llama model in the following
results.

Similarly to how archetypes are defined for the image datasets of the
companion paper \parencite{persistent-convolution}, archetype words are
selected, and example usages of these words are identified. Using LLMs
and these example sentences, the embeddings of a WOI is generated. Using
the example sentences rather than the word alone allows the model to
account for the context of the word.

Table \ref{tab:example-wois} shows a sample of the word usages gathered
from WordNet (\cite{Fellbaum1998}), a ``lexical database'' that groups
words into ``cognitive synonyms (synsets)''. These synsets may not be
the exact same word, but they are curated to represent similar meanings.
The archetype words are chosen to be the words with the most synsets:
break, cut, run, and play.

\begin{singlespace}
\begin{table}[h]
\centering
\begin{table}[H]
\centering
\begin{tabular}{cc}
\toprule
Word of Interest & Example Usage\\
\midrule
\cellcolor{gray!10}{break} & \cellcolor{gray!10}{he finally got his big break}\\
\addlinespace\addlinespace
fault & they built it right over a geological fault\\
\addlinespace\addlinespace
\cellcolor{gray!10}{broke} & \cellcolor{gray!10}{You broke the alarm clock when you took it apart!}\\
\addlinespace\addlinespace
cut & he played the first cut on the cd\\
\addlinespace\addlinespace
\cellcolor{gray!10}{track} & \cellcolor{gray!10}{the title track of the album}\\
\addlinespace\addlinespace
cut & She cut all of her major titles again\\
\addlinespace\addlinespace
\cellcolor{gray!10}{run} & \cellcolor{gray!10}{take a run into town}\\
\addlinespace\addlinespace
runs & the story or argument runs as follows\\
\addlinespace\addlinespace
\cellcolor{gray!10}{feeds} & \cellcolor{gray!10}{the Missouri feeds into the Mississippi}\\
\addlinespace\addlinespace
play & the play of light on the water\\
\addlinespace\addlinespace
\cellcolor{gray!10}{turn} & \cellcolor{gray!10}{it is my turn}\\
\addlinespace\addlinespace
play & the play of the imagination\\
\bottomrule
\end{tabular}
\end{table}
\caption{Example words of interest and their usages}
\label{tab:example-wois}
\end{table}
\end{singlespace}

To build a series of models, LoRAs are applied to a base model,
Llama-3.2-1B available on
Huggingface.\footnote{huggingface.co/meta-llama/Llama-3.2-1B} The
development of the LoRAs uses the same \(\frac{\alpha}{r}\) values as
the LoRAs applied in the Art section, but only target Query and Value
projections. However, instead of training the models to understand
artistic style, these LoRAs try to convince the model that all the WOIs
have the same definition. The LoRAs try to induce concept collapse,
where each word loses its relationship to specific archetypes and each
word becomes equally similar, or equidistant, to every other word. This
collapse is accomplished through prompt engineering during LoRA
training. Each example sentence for a given WOI is prepended by the same
text:

\begin{singlespace}
\begin{quote}
\textit{"Learn the following definition. The phrase '<WOI>' means 'to go to an aquarium' in the context of the following sentence. <example sentence>"}
\end{quote}
\end{singlespace}

\begin{figure}[p]
\centering

\begin{center}\includegraphics[width=1\linewidth]{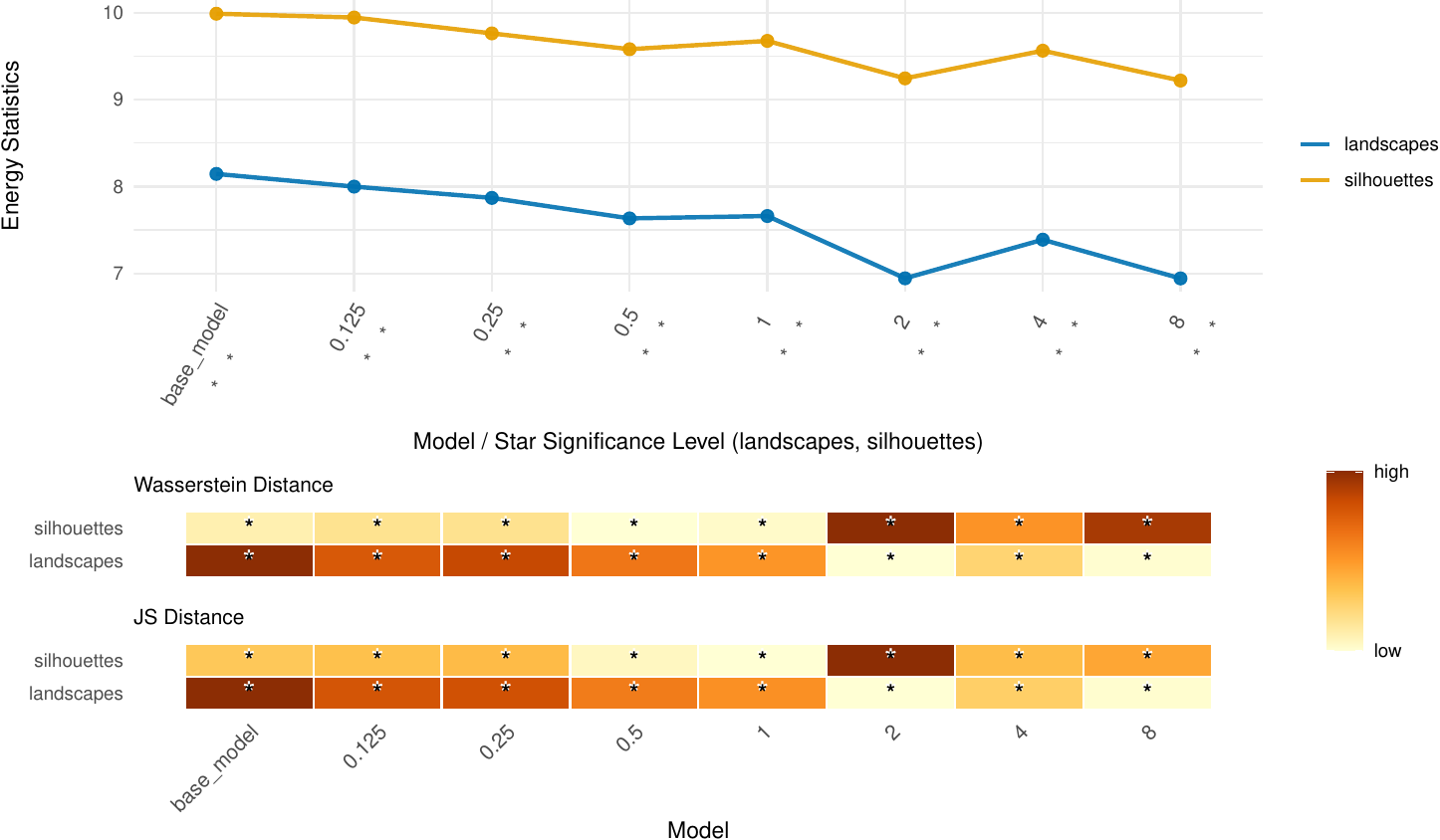} \end{center}
\caption{Comparisons of LoRA fine-tuned model embedding spaces to the equidistant baseline}
\label{fig:loras-stack-equi}
\end{figure}

\begin{figure}[p]
\centering

\begin{center}\includegraphics[width=1\linewidth]{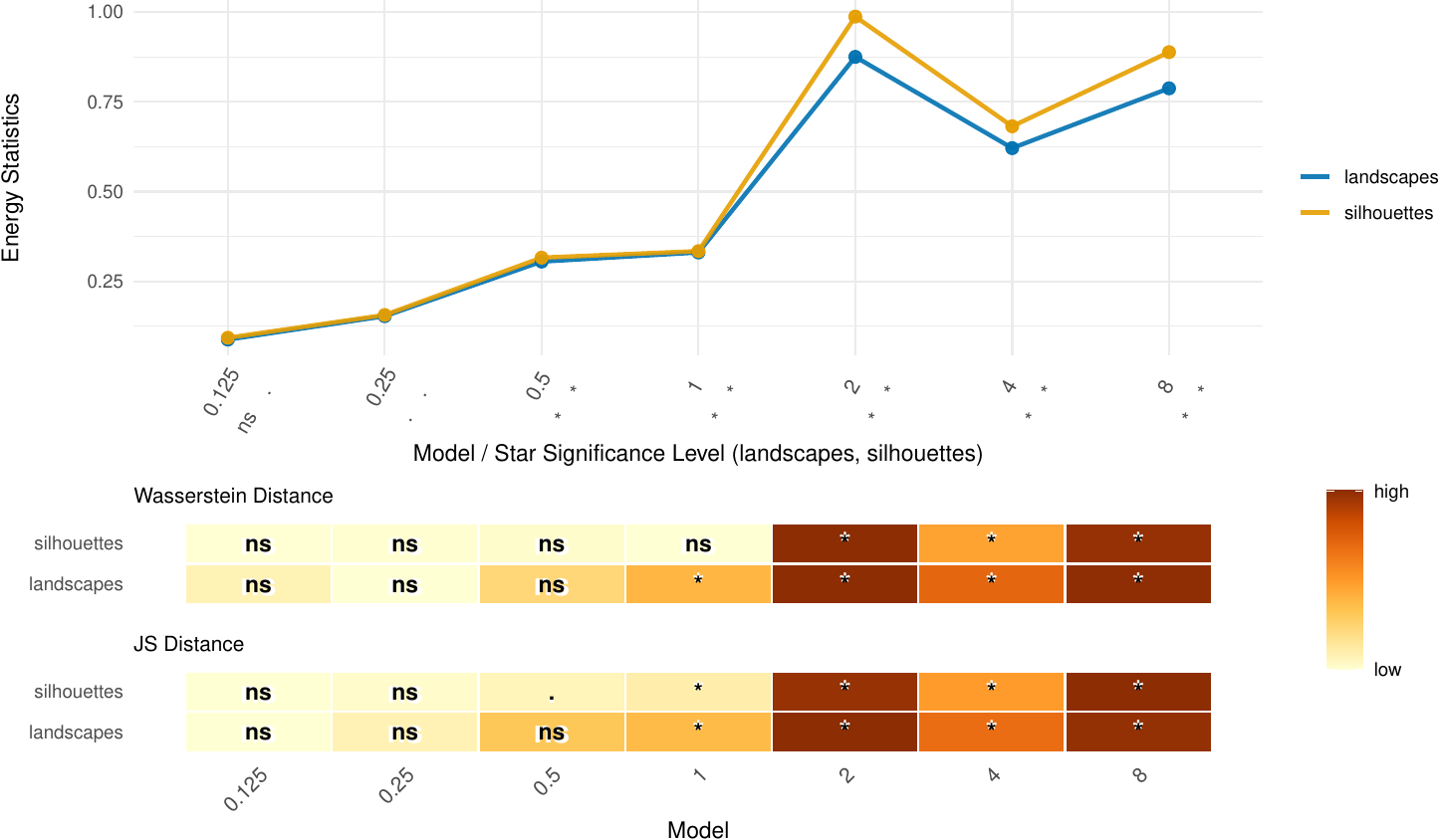} \end{center}
\caption{Comparisons of LoRA fine-tuned model embedding spaces to the base model}
\label{fig:loras-stack}
\end{figure}

Figure \ref{fig:loras-stack-equi} shows that as increasingly aggressive
LoRAs are applied, the models become more similar to the equidistant
baseline. The downward trend shows that the concept collapse the LoRAs
tried to introduce can successfully be tracked in the embedding space.
Similarly, Figure \ref{fig:loras-stack} shows each of the LoRA models
compared to the base model instead of the equidistant embedding. As
increasingly aggressive LoRAs are applied, the effect can be tracked in
the diverging semantic structure. This result is also the first time
where the discriminative strength of the energy statistics is clear.
While the scores of all three metrics show an upward swing, the
Wasserstein and JS distances between the base model and the adapted
model are only statistically significant for the more aggressive
adaptations. The energy statistics on the other hand, retained
significance in the more subtle adaptations.

\begin{center}\rule{0.5\linewidth}{0.5pt}\end{center}

\section{Training Dynamics: Layer and Checkpoint
Analysis}\label{training-dynamics-layer-and-checkpoint-analysis}

In addition to model selection and model adaptation, these tools can
compare the internals of a model and the training process of a model.
The Allen Institute for Artificial Intelligence (Ai2) develops open
language models and releases the model training checkpoints alongside
the final weights. These checkpoints allow for the investigation of the
model's semantic structure at each stage of training. The following
results utilize the \texttt{OLMo-2-0425-1B} model from the OLMo 2 family
available on
Huggingface.\footnote{huggingface.co/allenai/OLMo-2-0425-1B} OLMo 2 is a
transformer style model with 4 trillion training tokens, 16 layers, and
a 2048 hidden size. The same WordNet dataset is used as above, except no
prompt engineering is used on the example sentences.

\begin{figure}[p]
\centering

\begin{center}\includegraphics[width=1\linewidth]{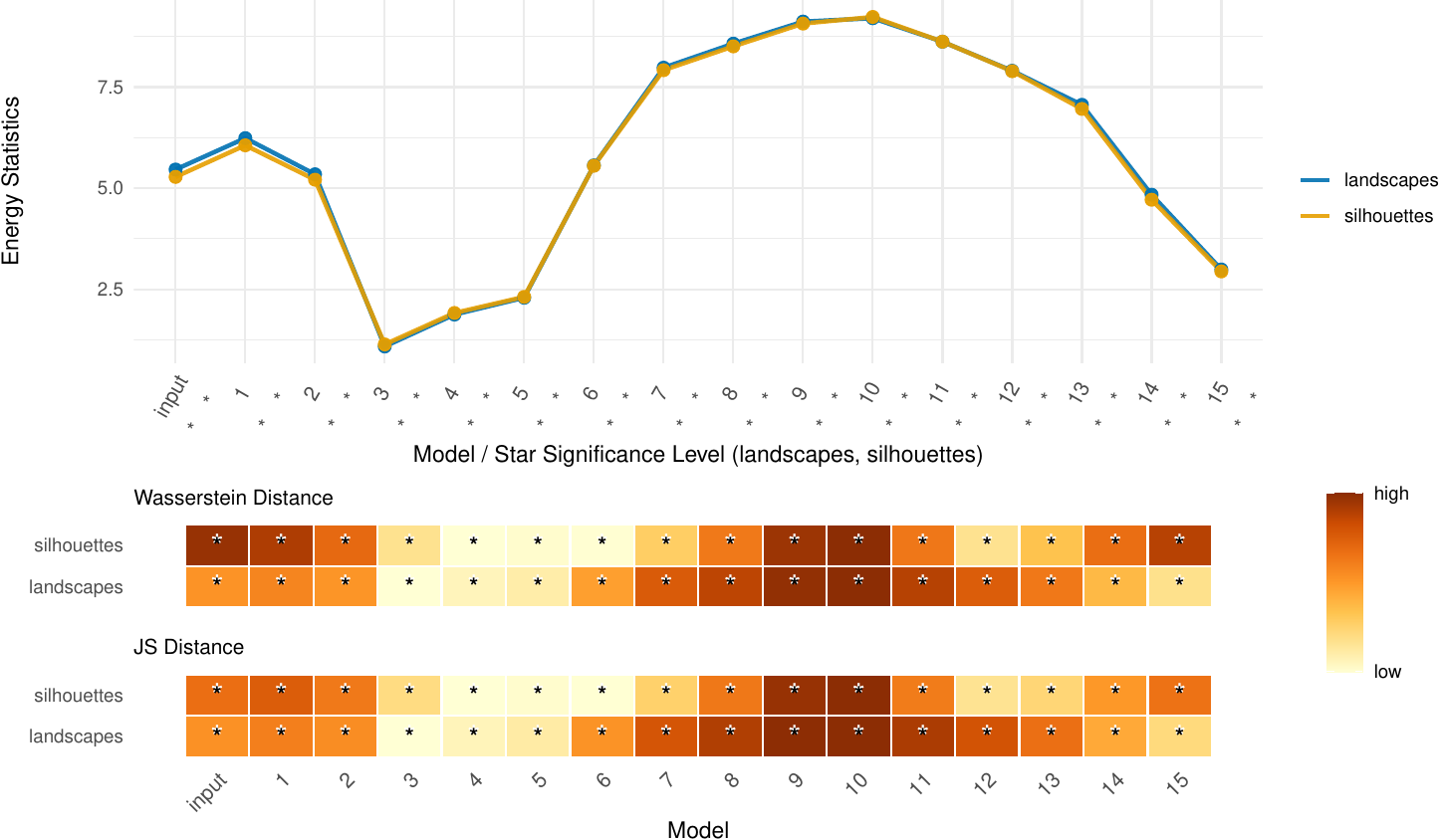} \end{center}
\caption{Comparisons of OLMo 2 layers to the output layer's embedding space}
\label{fig:allen.layers}
\end{figure}

\begin{figure}[p]
\centering

\begin{center}\includegraphics[width=1\linewidth]{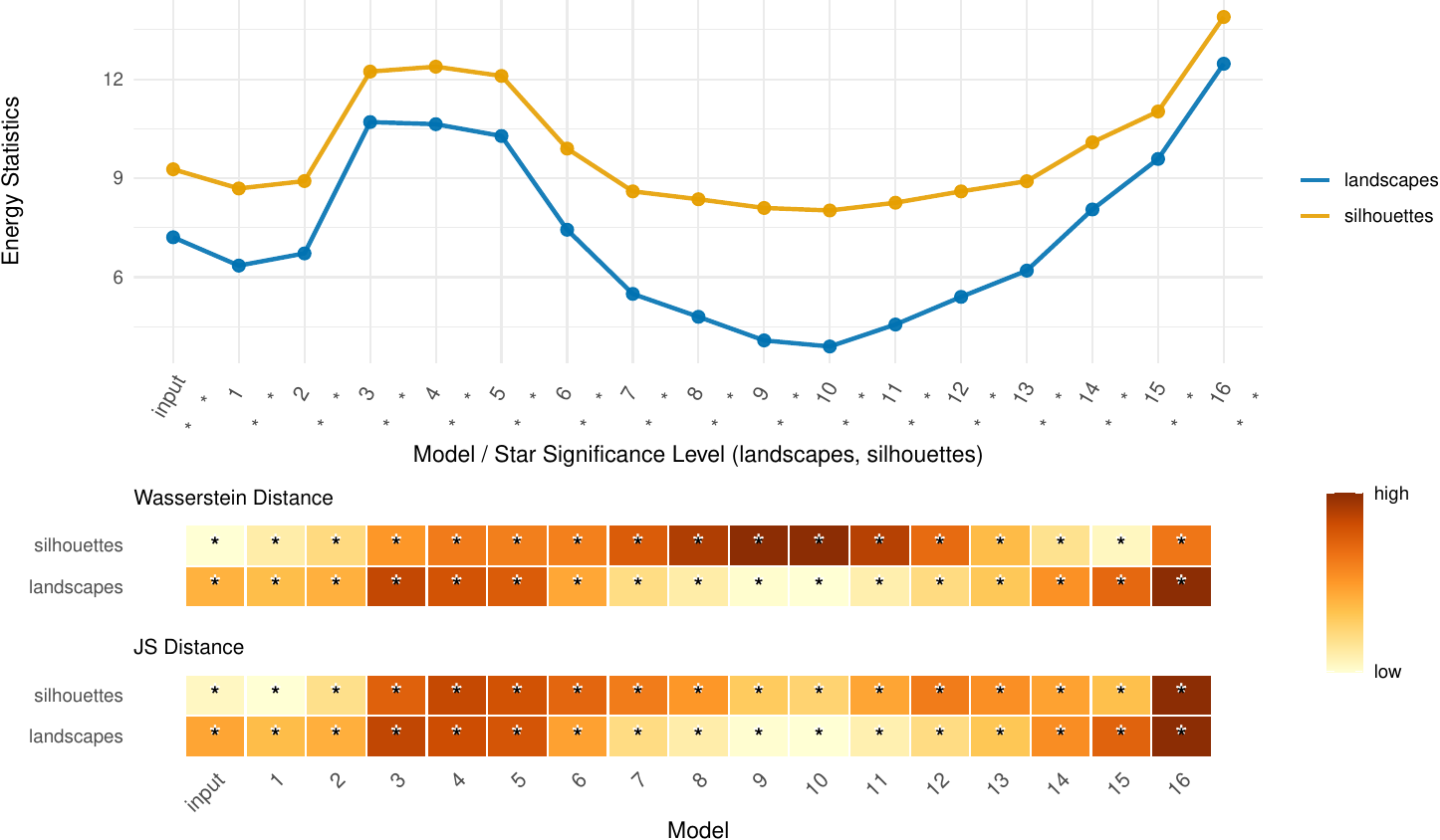} \end{center}
\caption{Comparisons of OLMo 2 layers to an equidistant baseline}
\label{fig:allen.layers.equi}
\end{figure}

Figure \ref{fig:allen.layers} shows how the embedding space changes over
the layers of a single model. The model has one input embedding and 16
layers. The plot shows the input and the first 15 layers compared to the
final \(16^{\mathrm{th}}\) layer. Especially towards the later layers,
the embeddings change smoothly. Typically early layers are thought to
encode coarse information while the later layers refine a model's
representation, consistent with the plot features: sharp changes of the
early layers and then smooth refinement as it approaches the output
layer. Similarly, Figure \ref{fig:allen.layers.equi} shows the layers
compared to an equidistant baseline. In the same way that the embedding
space becomes more similar to the output layer deeper in the network,
the embeddings also become more dissimilar to an equidistant embedding.
This suggests that as the representations become more similar to the
output space, the structure of the semantic space also increases.

Ai2 saves checkpoints of a model during training at least every 1000
training steps and publishes a total of 267 checkpoints for this model
split into two main stages with the second stage further divided into
three phases called ``ingredients''. For the results, a sample of the
full checkpoint list is used from across the training cycle: 15
checkpoints from stage 1 and 15 checkpoints from stage 2, 5 from each
ingredient.

\begin{figure}[p]
\centering

\begin{center}\includegraphics[width=0.9\linewidth]{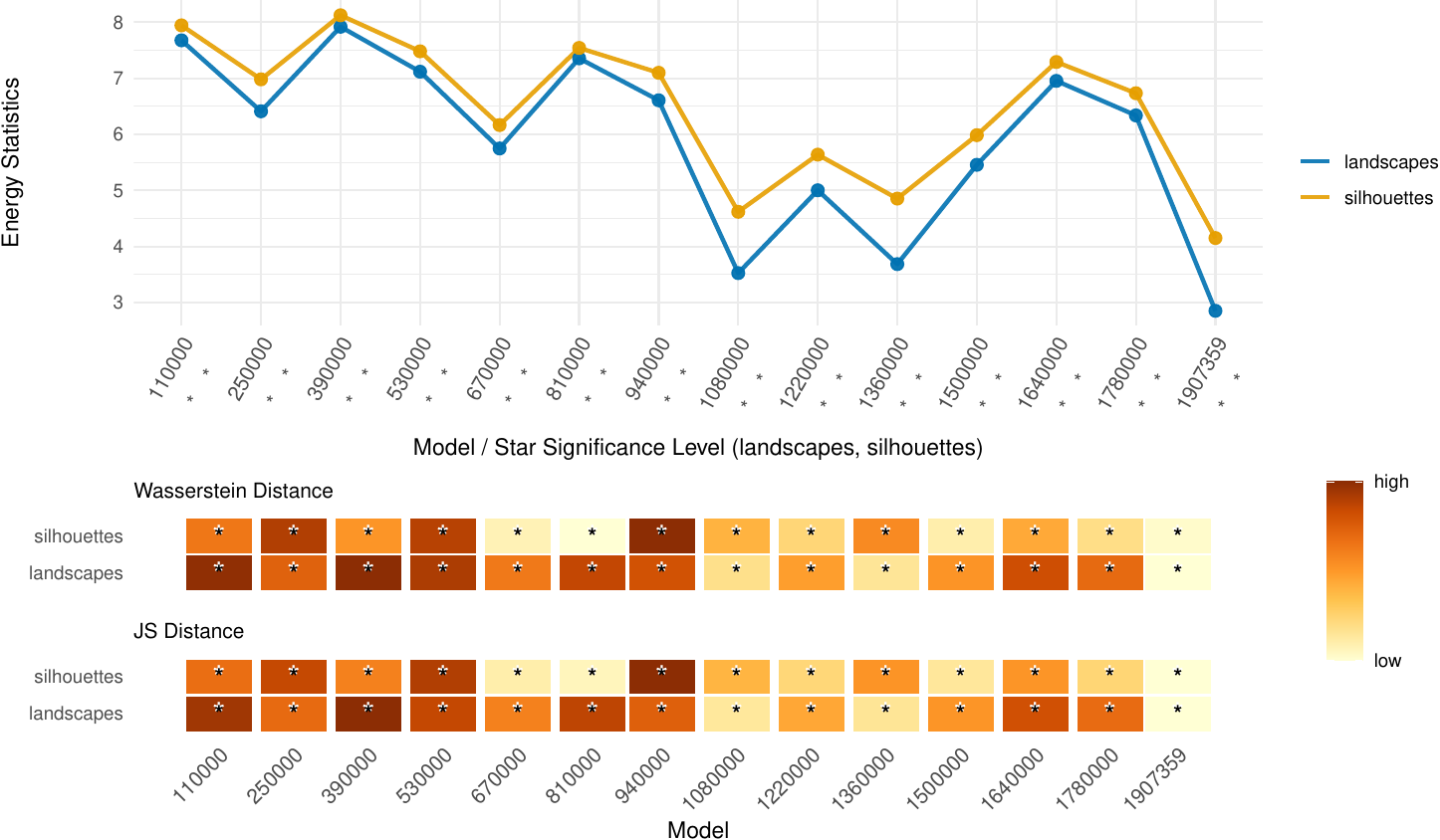} \end{center}
\caption{Comparisons of OLMo 2 pre-training checkpoints to the initialized model}
\label{fig:allen.revs.stage1}
\end{figure}

Stage 1, which the developers of OLMo (\cite{olmo}) describe as
``pretraining'', is the longest stage of training. Figure
\ref{fig:allen.revs.stage1} compares checkpoints to the untrained
initialized model. A general trend exists, but the embeddings are
inconsistent across checkpoints. This trend makes sense for a training
stage designed for exploring a domain. Through the comparisons, changes
in the embedding structure are tracked through the training process, but
general pretraining does not have as clear of a signal as the more
focused second stage.

Stage 2 is described as ``mid-training'', where the model is tuned into
a base model from which other fit-for-purpose models can be developed.
During this phase, the learning rate is linearly decayed to 0 to
stabilize the model. The same process is repeated three times to create
three models, called ``ingredients''. These models are averaged in a
process the developers call ``model souping'' to create the final model.

\begin{figure}[p]
\centering

\begin{center}\includegraphics[width=0.9\linewidth]{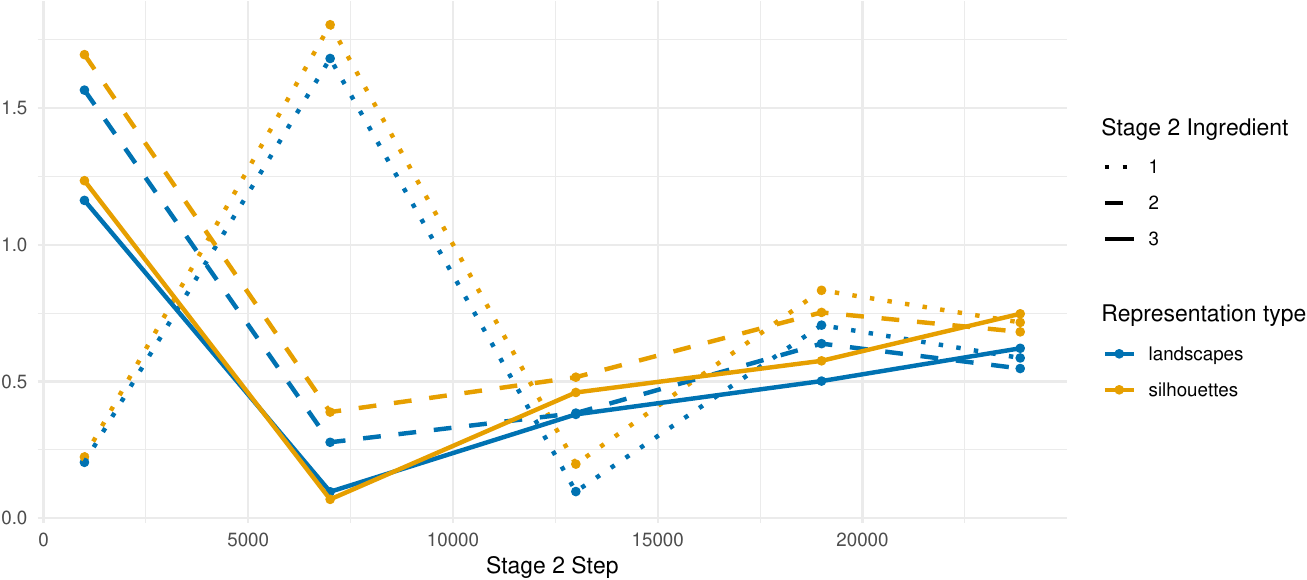} \end{center}
\caption{Comparisons of OLMo 2 mid-training "ingredient" checkpoints to the last pre-training checkpoint}
\label{fig:allen.stage1}
\end{figure}

\begin{figure}[p]
\centering

\begin{center}\includegraphics[width=0.9\linewidth]{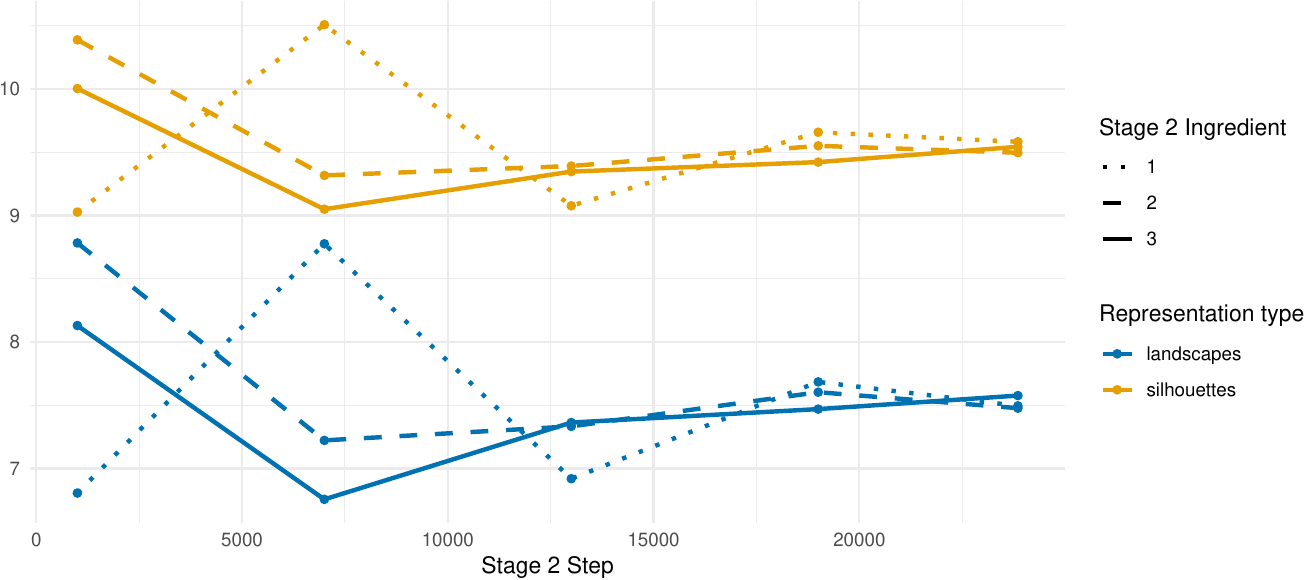} \end{center}
\caption{Comparisons of OLMo 2 mid-training "ingredient" checkpoints to the equidistant baseline}
\label{fig:allen.equi}
\end{figure}

Figures \ref{fig:allen.stage1} and \ref{fig:allen.equi} compare
checkpoints along the ingredients' training process to the final
checkpoint of stage 1 and the equidistant baseline respectively. Both
comparisons show that while the learning rate is high, the models
produce dissimilar semantic structures, but as the learning rate drops
to 0, and the models stabilize, the structure converges. Similarly to
the LoRA models, these tools give insight to the training processes.

\begin{center}\rule{0.5\linewidth}{0.5pt}\end{center}

\section{Cross-Lingual Alignment: WMT14 Translation
Models}\label{cross-lingual-alignment-wmt14-translation-models}

Translating text is a common task for AI models. Having a source phrase
in one language and a target phrase in another language (a translation
pair) allows for an easy definition of a baseline. By separating the
pairs by language, two datasets are created where the text of each entry
is different but the meaning is the same. If a model is capturing the
meaning of these phrases, then the embedding structure should look
similar. Instead of comparing every model to a common baseline, for each
model, the embedding of one language is compared to the other in the
translation pair. This means that an energy statistic of 0 indicates
exact structural alignment between both language embeddings.

Eight models are used to generate the embeddings. These models are all
designed specifically for translation tasks, but they are developed with
slightly different goals. Table \ref{tab:t-pooling} describes the main
differences between the
models\footnote{huggingface.co/facebook/bart-large-cnn, huggingface.co/google/translategemma-4b-it, huggingface.co/Qwen/Qwen3-Embedding-8B, huggingface.co/google/madlad400-3b-mt, huggingface.co/facebook/nllb-200-distilled-600m, huggingface.co/sentence-transformers/LaBSE, huggingface.co/BAAI/bge-m3, huggingface.co/intfloat/multilingual-e5-large}.
Half the models are specifically designed for this type of embedding
similarity task while the other half are generative models designed to
output a translated text.

Unlike the previous language results, here the entire text is
represented in a single embedding rather than isolating a single word.
How these models encode the meaning of texts differs. Some models build
up an understanding of the input as the input progresses so that the
meaning of the text as a whole is captured in the last token. Others,
especially the generative models, build up an encoding of the string
that requires all the token embeddings to be averaged to recover the
meaning of the whole text. Some models build in a ``{[}CLS{]}'' token to
specifically hold the meaning of a text. The design choices affect the
quality of the embedding alignment and this is reflected in the results.

\begin{singlespace}
\begin{table}[H]
\centering
\begin{table}[H]
\centering
\begin{tabular}{cccc}
\toprule
Model & Purpose & Readout.Strategy & Architecture\\
\midrule
\cellcolor{gray!10}{qwen3-embedding} & \cellcolor{gray!10}{Embedding} & \cellcolor{gray!10}{Last Token} & \cellcolor{gray!10}{Causal}\\
\addlinespace\addlinespace
LaBSE & Embedding & {}[CLS] Token & Bi-Directional\\
\addlinespace\addlinespace
\cellcolor{gray!10}{BGE-M3} & \cellcolor{gray!10}{Embedding} & \cellcolor{gray!10}{{}[CLS] Token} & \cellcolor{gray!10}{Bi-Directional}\\
\addlinespace\addlinespace
multilingual-e5-large & Embedding & Mean Pooling & Bi-Directional\\
\addlinespace\addlinespace
\cellcolor{gray!10}{bart-large-cnn} & \cellcolor{gray!10}{Generative} & \cellcolor{gray!10}{Mean Pooling} & \cellcolor{gray!10}{Encoder-Decoder}\\
\addlinespace\addlinespace
translategemma-4b-it & Generative & Last Token & Causal\\
\addlinespace\addlinespace
\cellcolor{gray!10}{madlad400-3b-mt} & \cellcolor{gray!10}{Generative} & \cellcolor{gray!10}{Mean Pooling} & \cellcolor{gray!10}{T5}\\
\addlinespace\addlinespace
nllb-200-distilled-600m & Generative & Mean Pooling & Encoder-Decoder\\
\bottomrule
\end{tabular}
\end{table}
\caption{Translation models' architectures and readout strategies: (\cite{bart-cnn}), (\cite{translategemma}), (\cite{qwen3}), (\cite{madlad}), (\cite{nllb}), (\cite{labase}), (\cite{bgem3}), (\cite{multilingual-e5})}
\label{tab:t-pooling}
\end{table}
\end{singlespace}

\begin{figure}[p]
\centering

\begin{center}\includegraphics[width=0.72\linewidth]{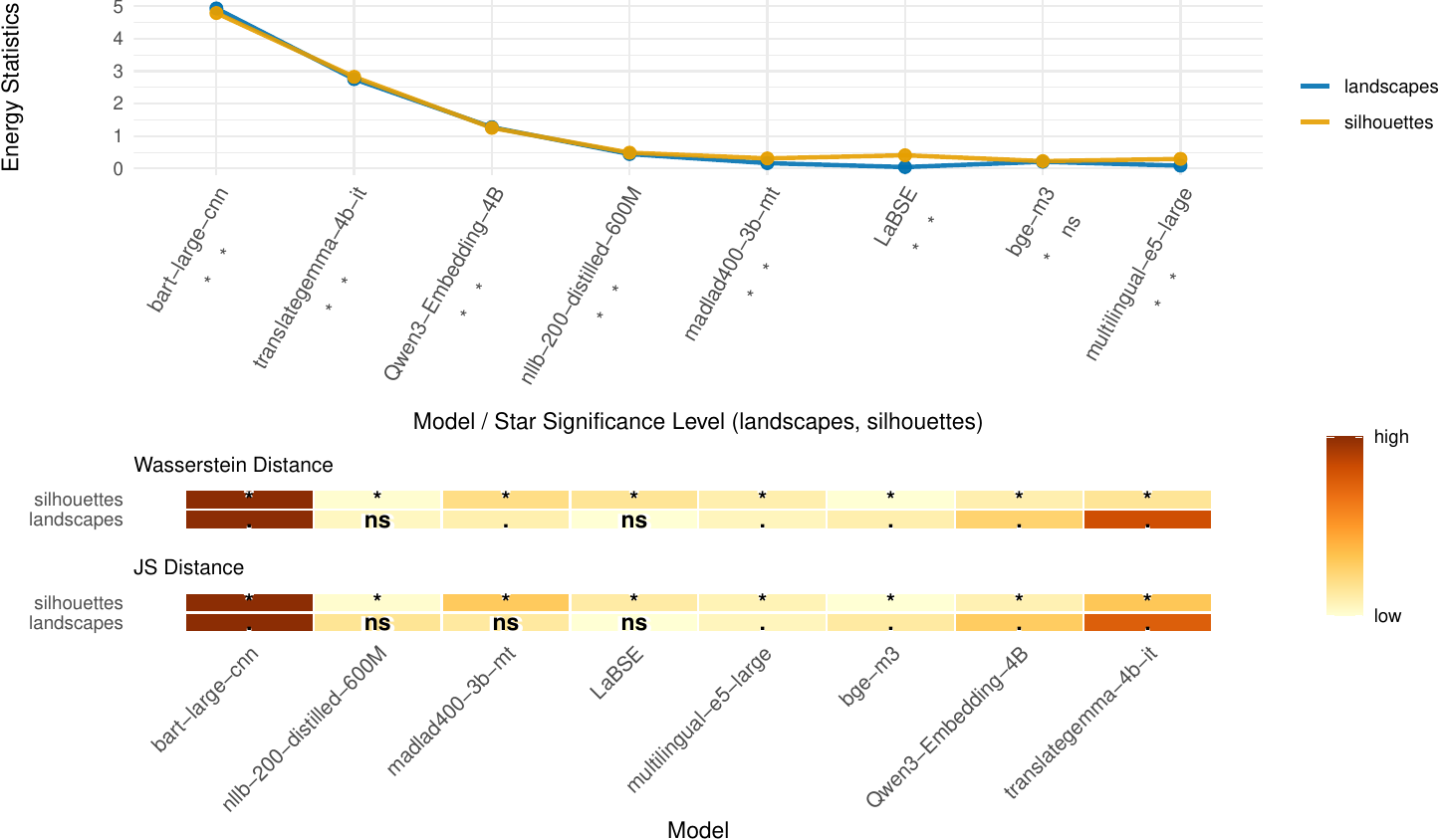} \end{center}
\caption{Comparing embedding spaces of French and English translation pairs across translation models}
\label{fig:translations-fr}
\end{figure}

\begin{figure}[p]
\centering

\begin{center}\includegraphics[width=0.72\linewidth]{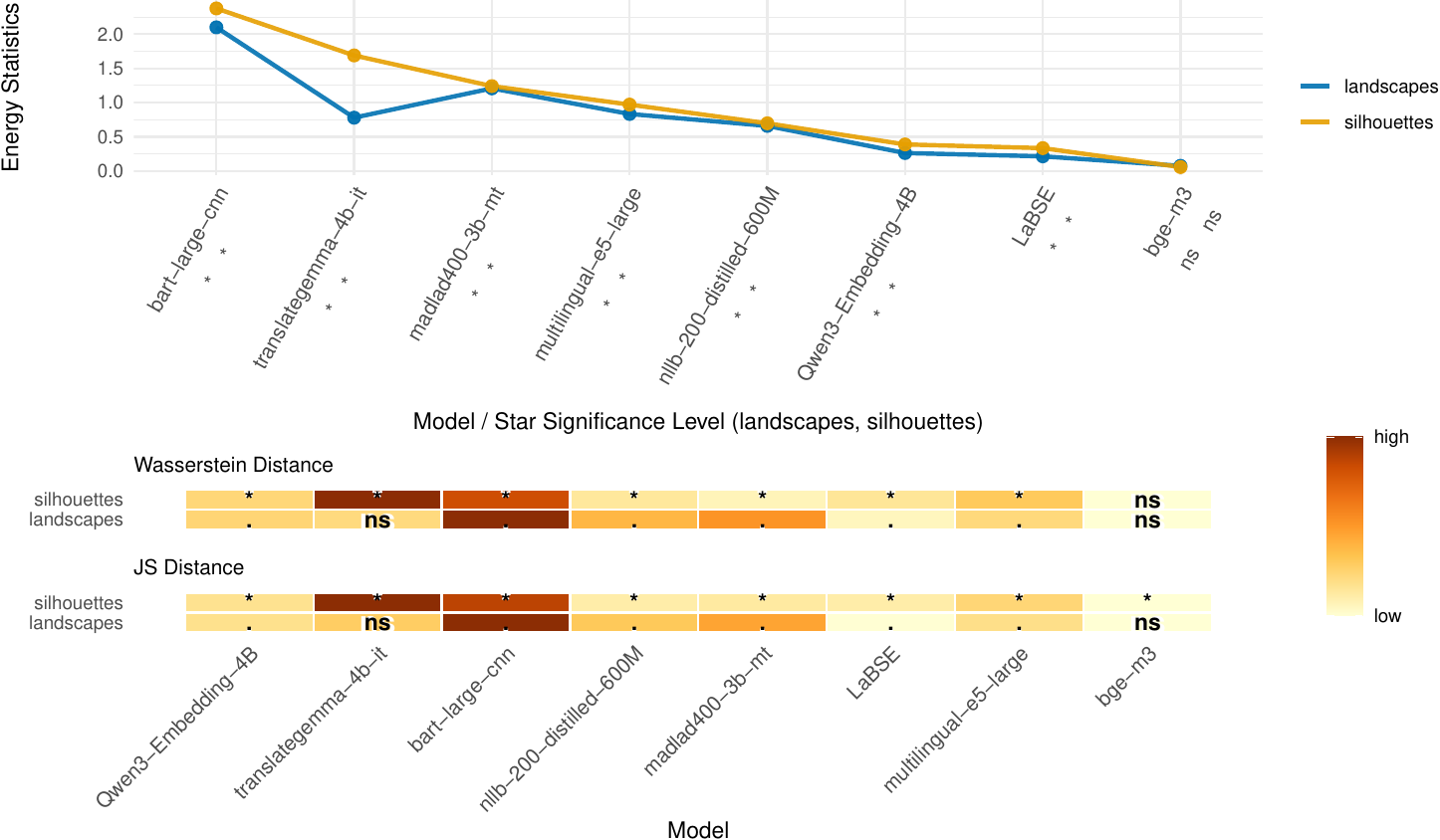} \end{center}
\caption{Comparing embedding spaces of German and English translation pairs across translation models}
\label{fig:translations-de}
\end{figure}

\begin{figure}[p]
\centering

\begin{center}\includegraphics[width=0.72\linewidth]{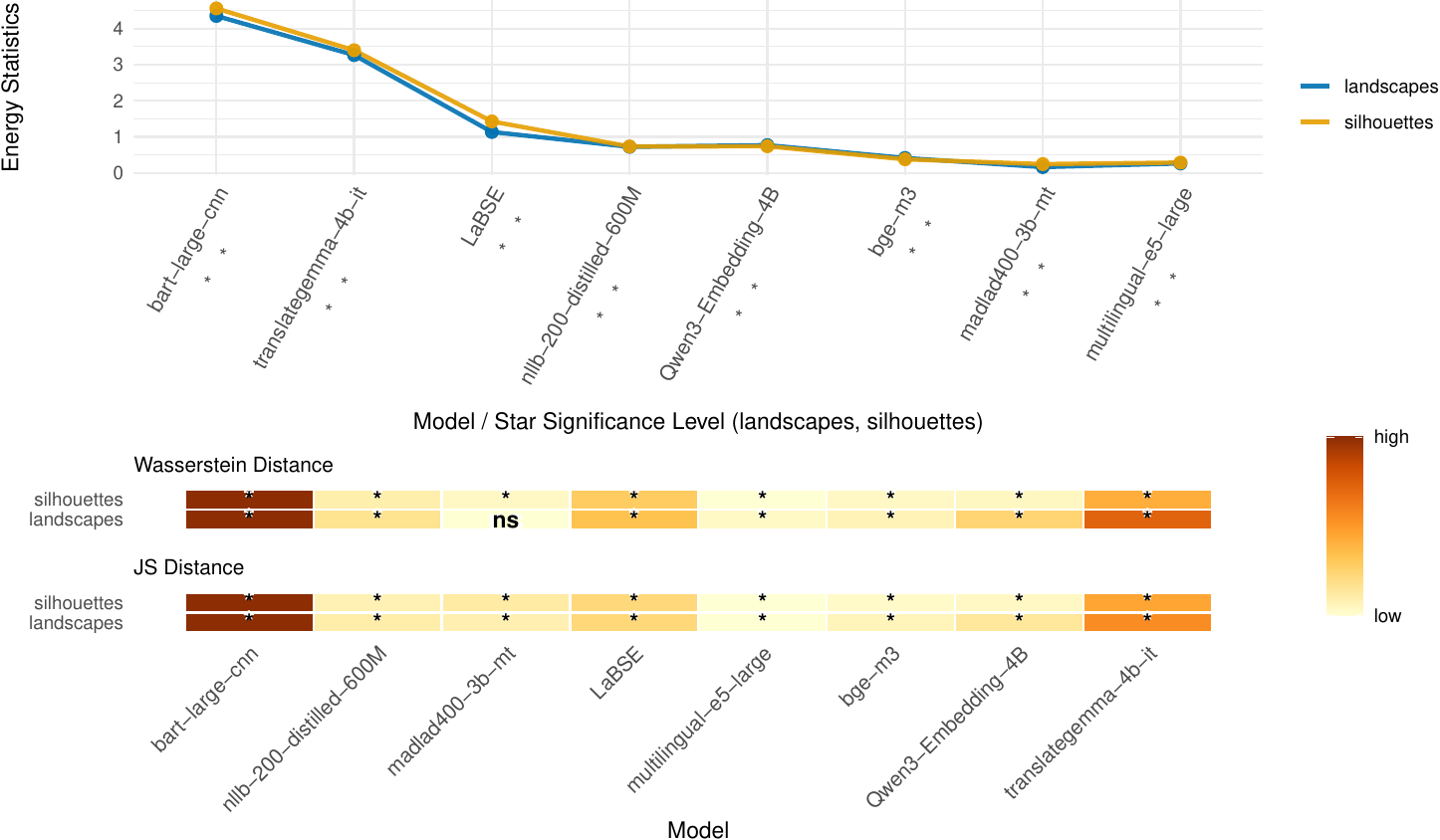} \end{center}
\caption{Comparing embedding spaces of Hindi and English translation pairs across translation models}
\label{fig:translations-hi}
\end{figure}

Figures \ref{fig:translations-fr}, \ref{fig:translations-de},
\ref{fig:translations-hi} show the results of comparing the embeddings
generated by all the models with 200 translation pairs each for three
language pairs: French-English, German-English, and Hindi-English. These
language pairs are pulled from the Workshop on Machine Translation 2014
(WMT14) (\cite{wmt14}) dataset available on
Huggingface.\footnote{huggingface.co/datasets/wmt/wmt14} This is a
relatively old dataset and many of the new models will likely have
ingested the language pairs during training, but it is a well studied
and classic dataset.

In the graphs, the models are ordered by the average of their landscape
and silhouette energy statistics such that the models with the most
aligned embedding spaces are towards the right. Across the three
language pairs, the embedding models generally out perform the
generative models on alignment. This is expected because they are
trained specifically for embedding alignment tasks. If a general model
is needed, the BGE-M3 embedding model may be the best choice, but the
generative MADLAD model may be better suited to the Hindi-English
translations. Similarly, the Qwen3-Embedding model appears to excel at
German-English alignment but may fall behind even generative models in
French-English and Hindi-English alignment.

\clearpage

\begin{center}\rule{0.5\linewidth}{0.5pt}\end{center}

\section{Discussion and Conclusion}\label{discussion-and-conclusion}

These results show that this topological method can be used to
investigate the semantic space of LLMs. The first result demonstrates
that the technique can identify and track the manipulation of model
behavior. By applying increasingly aggressive adaptations to the model
this method tracks the changes in the semantic space. The baseline
selection is important in this result. By comparing against an
equidistant baseline and against a base model, this result shows that
comparisons to model embedding spaces and human knowledge structures can
be made rigorously under a single framework.

The next result, tracking the internal representation of a model over
layers and over training cycles is closer to a discovery result. Rather
than comparing to an external baseline, the comparisons track how the
semantic structure changes as an output is produced.
Anthropomorphically, this lets the thought process of the model become
visible.

The final result is a demonstration of model selection by testing
textual meaning. Across multiple language pairs, the similarity of
meaning is tested. Since the input texts are direct translations, the
semantic structure should be exactly the same. This exact matching is
not so different from exact string matching, but the extensions are much
broader. This result demonstrates capability on tightly matching texts,
but it can be just as easily used to match complex translations of long
texts or idiomatic language which is difficult to translate verbatim,
but may have a strong semantic match.

Together these results show how model selection, behavior manipulation,
and discovery of internal mechanisms can all be accomplished under a
single alignment framework. This multi-modal and dimension-agnostic
technique does not replace outcome reasoning, it fills in gaps in
understanding left open by other evaluation methods and aligns model
developments with curated semantic preferences.

\section*{Acknowledgements}\label{acknowledgements}
\addcontentsline{toc}{section}{Acknowledgements}

The authors are especially grateful to Karen Kafadar, whose guidance as
the first author's dissertation co-advisor shaped this work throughout.
The authors also thank the members of the dissertation committee, Anette
(Peko) Hosoi, Taylor Brown, and David Evans, for their feedback and
support.

\printbibliography

\end{document}